\documentclass{article}

\usepackage{arxiv}

\usepackage[utf8]{inputenc} 
\usepackage[T1]{fontenc}    
\usepackage{hyperref}       
\usepackage{url}            
\usepackage{booktabs}       
\usepackage{amsfonts}       
\usepackage{nicefrac}       
\usepackage{microtype}      
\usepackage{graphicx}
\usepackage{natbib}
\usepackage{doi}

\usepackage{multirow}
\usepackage[normalem]{ulem}
\newcommand{\tb}[1]{\textbf{#1}}
\newcommand{\dataset}{{\textit{LingFeatured NLI}}}
\newcounter{example}
\newenvironment{example}[1][]{
\refstepcounter{example}
   \par\medskip
   \noindent \textbf{<T, H> Example~\theexample. #1} \rmfamily}{\medskip}

\title{Polish Natural Language Inference and Factivity -- an Expert-based Dataset and Benchmarks}

\author{ \href{https://orcid.org/0000-0003-0506-4751}{\includegraphics[scale=0.06]{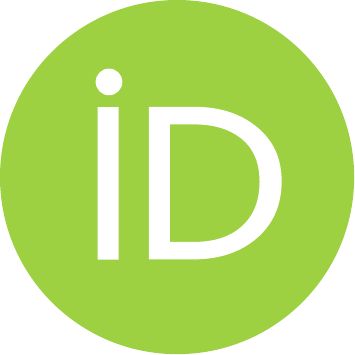}\hspace{1mm}Daniel Ziembicki} \\
    Department of Formal Linguistics,\\
        University of Warsaw, Warsaw, Poland \\
        \texttt{daniel.ziembicki@uw.edu.pl}\\
	\And
	\href{https://orcid.org/0000-0002-3407-7570}{\includegraphics[scale=0.06]{orcid.pdf}\hspace{1mm}Anna Wr{\' o}blewska} \\
	Faculty of Mathematics and Information Science,\\
	Warsaw University of Technology,\\
	Warsaw, Poland \\
	\texttt{anna.wroblewska1@pw.edu.pl} \\
	\And
	\href{https://orcid.org/0000-0003-0617-7301}{\includegraphics[scale=0.06]{orcid.pdf}\hspace{1mm}Karolina Seweryn} \\
	Faculty of Mathematics and Information Science,\\
	Warsaw University of Technology,\\
	Warsaw, Poland \\
	\texttt{karolina.seweryn@pw.edu.pl} \\
}

\renewcommand{\shorttitle}{Polish NLI and Factivity}

\hypersetup{
pdftitle={Polish Natural Language Inference and Factivity -- an Expert-based Dataset and Benchmarks},
pdfsubject={q-bio.NC, q-bio.QM},
pdfauthor={Daniel~Ziembicki, Anna~Wr{\' o}blewska, Karolina~Seweryn},
pdfkeywords={Natural Language Inference, Factivity, HerBERT, Low-resource Languages},
}

\begin{document}
\maketitle

\begin{abstract}

Despite recent breakthroughs in Machine Learning for Natural Language Processing, the Natural Language Inference (NLI) problems still constitute a challenge. 
To this purpose we contribute a new dataset that focuses exclusively on the factivity phenomenon; however, our task remains the same as other NLI tasks, i.e. prediction of entailment, contradiction or neutral (ECN). The dataset contains entirely natural language utterances in Polish and gathers 2,432 verb-complement pairs and 309 unique verbs. The dataset is based on the National Corpus of Polish (NKJP) and is a representative sample in regards to frequency of main verbs and other linguistic features (e.g. occurrence of internal negation).We found that transformer BERT-based models working on sentences obtained relatively good results ($\approx89\%$ F1 score). Even though better results were achieved using linguistic features ($\approx91\%$ F1 score), this model requires more human labour (humans in the loop) because features were prepared manually by expert linguists. BERT-based models consuming only the input sentences show that they capture most of the complexity of NLI/factivity. Complex cases in the phenomenon - e.g. cases with entitlement (E) and non-factive verbs - remain an open issue for further research.

\end{abstract}

\keywords{Natural Language Inference \and Factivity \and HerBERT, Low-resource Languages}

\section{Introduction}
Semantics is still one of the biggest problems of Natural Language Processing (NLP)\footnote{See relatively new and important work related to this topic: \citep{richardson2020probing, zhang2020semantics}.}. It should not come as a surprise; semantic problems are also the most challenging in the field of linguistics itself (see~\citep{sep-meaning}). The topic of presupposition and such relations as entailment, contradiction and neutrality are at the core of semantic-pragmatic research~\citep{huang201114}. For this reason, we dealt with the issue of factivity, which is one of the types of presupposition.

The subject of this study includes three phenomena occurring in the Polish language. The first of them is \textbf{factivity} \citep{Kiparsky-1971, karttunen1971some}. The next phenomenon is the three relations: \textbf{entailment}, \textbf{contradiction} and \textbf{neutrality} (ECN), often studied in Natural Language Inference (NLI) tasks. The third and last phenomenon is utterances with the following syntactic pattern: '\textbf{[verb][że][complement clause]}'. The segment \emph{że} corresponds to the English segments \emph{that} and \emph{to}.

This study aims to answer the following research questions (RQs): 
\begin{enumerate}
    \item[RQ1:] How relevant is the opposition factivity / non-factivity for the prediction of ECN relations?
    \item[RQ2:] How well do machine learning (ML) models recognize the entailment, contradiction and neutrality in utterances with the structure '[verb][that][complement clause]'? 
\end{enumerate}

To answer the first question, we collected a dataset based on the National Corpus of Polish (NKJP). The NKJP corpus is the largest Polish corpus which is genre diverse, morphosyntactically annotated and representative of contemporary Polish~\citep{Przepiorkowski:12}. Our goal was to prepare a dataset that is representative and adequately reflects the problems in NLI in the Polish language. Additionally, besides the utterances, we prepared multiple linguistic features to study and compare the influence on our main task -- prediction of ECN relations. 

The answer to the second question required building ML models. We trained models based on prepared linguistic features and text embedding BERT variants.
Note that if a presupposition trigger in the form of a factive verb is used in an utterance, then -- under certain conditions -- an entailment relation occurs between the whole sentence and its complement. If a non-factive verb is used in an utterance, one of the following three relations is possible: entailment, contradiction, or neutrality. We investigated whether the modern machine learning models handle this linguistic phenomenon.

We adopted the following research perspective: first, we chose a linguistic issue, namely the opposition of factivity vs non-factivity. The issue of factivity has posed theoretical problems since this phenomenon was first noticed. Factivity has also not been a subject of research for the Polish language so far. Then, we created a dataset that reflects this opposition in communication conducted in Polish. With a dataset, we posed the research question outlined above. It is worth mentioning that we adopted a different procedure than those commonly used in ML papers. Usually, researchers gather datasets regarding the chosen ML task (e.g., classification task, sentiment analysis, predicting ECN), which is sometimes not even related to linguistic problems. In this standard ML approach, the datasets are usually as big as possible and sometimes even synthetically generated or specially prepared by linguists. This is also different in our procedure, in which we chose only natural utterances that appeared in authentic communication (from NKJP corpus).

\noindent Thus, in this paper, our contributions are as follows:
\begin{itemize} 
\item gathering a new dataset \dataset, based on fully natural utterances from (NKJP). The dataset consists of 2,432 'verb-complement' pairs. It was enriched with various linguistic features to perform inferences of the utterance relation types, i.e. entailment, contradiction, neutral \tb{(ECN)} (see Section~\ref{sec:dataset}). To the best of our knowledge, it is the first such dataset in the Polish language. Additionally, all the utterances constituting the dataset were translated into English.
\item building ML benchmark models (linguistic feature-based and embedding-based BERT) that predict the utterance relation type \tb{ECN} (see Section~\ref{sec:ML-modeling}).\footnote{The dataset, its translation and the source code, and our ML models are attached as supplementary material and will be publicly available after acceptance.}
\end{itemize}

In the following, Section~\ref{sec:related_work} and Section~\ref{sec:related-datasets} describe theoretical background, related work and datasets. Then, Section~\ref{sec:dataset} introduces our new dataset \dataset with a discussion about [with a commentary on?] its language background, annotation process and features.  Section~\ref{sec:ML-modeling} shows our machine learning modelling approach and experiments. Further, in Section~\ref{sec:Results Analysis/Discussion}, we analyze the results and formulate findings. Finally, we summarize our work and indicate its main findings and limitations in~Section~\ref{sec:conclusions}.

\section{Linguistic Background}
\label{sec:related_work}

\subsection{Linguistic Problems}
In the linguistic and philosophical literature, the topic of factivity is one of the most disputed. The work of Kiparsky and Kiparsky~\citep{Kiparsky-1971} began the never-ending debate about presupposition and factivity in linguistics. This topic was of interest to linguists especially in the 1970s (see, e.g., \citep{karttunen1971some,givon1973time,elliott1974toward, hooper1975assertive, delacruz1976factives, stalnaker1977pragmatic}). 
Since the entry of the term  \emph{factivity} into the linguistic vocabulary in the 1970s, there have been many, often mutually exclusive, theoretical studies of this phenomenon.

Karttunen's~\citep{karttunen2016presupposition} article with the significant title \emph{Presupposition: What went wrong?} emphasized this fact. He mentioned that factivity is the research area that raised the most controversy among linguists. It should be clearly stated that no single dominant concept explaining the phenomenon of factivity has emerged so far. Nowadays, new research on this topic appears constantly, e.g., \citep{giannakidou2006only,egre2008question,beaver20103,tsohatzidis2012forget,anand2014factivity,kastner2015factivity,djarv2019factive}. The clearest line of conflict concerns the question in which area to situate the topic of factivity: semantics or pragmatics. There is also a dispute about discursive concepts (see, e.g., \citep{abrusan2016presupposition, tonhauser2016prosodic, simons2017best}).  An interesting example from our research point of view is the work of~\citep{djarv2020prosodic}. These authors argue against the claim that factivity depends on the prosodic features of the utterance, pointing out that it is a lexical rather than a discursive phenomenon.

In addition to the disputes in the field of linguistics, there is also work (see, e.g.,  \citep{hazlett2010myth, hazlett2012factive}) of a more philosophical orientation which strikes at beliefs that are mostly considered accurate in linguistics, e.g., that  \emph{know that} is a factive verb. These works, however, have been met with a distinctly polemical response (see, for example, \citep{turri2011mythology} \citep{tsohatzidis2012forget}).

In summary, \textbf{the first problem} to note is the clear differences in theoretical discussions of the phenomenon of factivity-based presupposition. These take place not only in the older literature, but also in the contemporary one.

Theoretical differences are related to \textbf{another issue}, namely the widespread disagreement about which verbs are factive and which are not. A good example is a verb \emph{regret that}, which, depending on the author, is factive or not or presents a different type of factivity from the paradigmatic in the class of factive expressions verb \emph{know that}\footnote{See on the one hand the works of \citep{karttunen1971some}, \citep{ozturk2017corpus}, \citep{dietz2018reasons}, on the other hand \citep{egre2008question}, \citep{dahlman2016did}, \citep{grigore2016factive}.}. 

The explosion of work on presupposition in the 1970s and the multiplicity of theoretical concepts resulted in the uncontrolled growth of terminological proposals and changes in the meaning of terms already used. The term \emph{presupposition} has been ambiguous since the 1970s, and this state of matters persists today. Terms such as \emph{factivity}, \emph{presupposition},  \emph{modality} or \emph{implicature} are indicated as typical examples of ambiguous expressions. Problems of terminology are \textbf{the third problem} to be highlighted in the work on factivity. It is important to note the disturbing phenomenon of transferring terminological issues to the NLI. Reporting studies analogous to ours will bring attention to this difficulty.

\textbf{A final point} to note is the lack of linguistic papers that provide a fairly exhaustive list of factive, non-factive, veridical, etc. expressions. There is also a lack of comparative work between ethnic languages in general. This kind of research is only available for individual, selective expressions (see e.g. \citep{ozyildiz2017factivity, hanink2017factivity, jarrah2019factivity, dahlman2019testing}).

\subsection{Key Terms}
\label{sec:terms}
We, therefore, chose to put more emphasis on conceptual issues. The concepts most important to this study will now be presented, primarily factive presupposition.

\subsubsection{Entailment, Contradiction, Neutral}

Let's start by introducing an understanding of the basic semantic relations:
\begin{itemize}
    \item \textbf{Entailment}: \textbf{H} must be true if \textbf{T} is true;
    \item \textbf{Contradiction}: \textbf{H} must be false, if \textbf{T} is true;
    \item \textbf{Neutral}: \textbf{H} may be false or true if \textbf{T} is true \\
    ~\citep{chierchia2000meaning}.
\end{itemize}  
\noindent where (\tb{T}) is an utterance and (\tb{H}) is an item of information.

\subsubsection{Information}
The information we are interested in is that transmitted by means of spoken sentences, which are utterances. The mechanisms involved in this transmission may be either of a \textit{purely semantic} (lexical consequences) or \textit{pragmatic nature} (conversational/scalar implicatures) \citep{grice1975logic,sauerland2004scalar}. Three examples of information are shown below.
\begin{example} \label{ex:ex1}
 (Entailment) \\ 
  \tb{T}: Only then did they realize that they had been cut off from their only way out. \\
  \tb{H}: They had been cut off from their only way out.  
\end{example}
\begin{example} \label{ex:ex2}
(Contradiction) \\
\tb{T}: If it wasn’t for the special smell of medicines in the air, you’d think it was just
a normal room.  \\
\tb{H}: It was just a normal room. 
\end{example}
\begin{example} \label{ex:ex3}
(Neutral) \\
\tb{T}: Statists believe that people can be demanding towards the state. \\
\tb{H}: People can be demanding towards the state.
\end{example}

In example ~\ref{ex:ex1}, the entailment is lexical in nature because it is founded on the factive verb \emph{realize that}. The nature of the contradiction in example ~\ref{ex:ex2} is not obvious; certainly (\tb{H}) is not a factive presupposition, since the verb \emph{think that} is not a factive unit. Regardless of the kind of contradiction we are dealing with here, there is just this relation between (\tb{T}) and (\tb{H}). In Example ~\ref{ex:ex3}, on the other hand, we have the neutrality: there is nothing in textbf{(T's)} utterance that guarantees either a lexical entailment or a contradiction, and there are no pragmatic mechanisms that would fund either of these two relations.

\subsubsection{Negation}
We take into account in our dataset the occurrence of negation, specifically internal negation. We distinguish it from the so-called external negation, which is not relevant to the phenomenon of factivity. The examples of these two types of negation can be found below. 
\begin{example} \label{ex:ex4} (Entailment) \\
\tb{T}: He didn't manage to open the door.\\
\tb{H}: He tried to open the door. 
\end{example}
\begin{example} \label{ex:ex5} (Neutral) \\
\tb{T}: It is not the case that he managed to open the door. \\
\tb{H}: He tried to open the door.
\end{example}

In Example~\ref{ex:ex4}, internal negation was used, and in Example~\ref{ex:ex5}, external negation. The utterance in~\ref{ex:ex4} implies (\tb{H}), whereas in~\ref{ex:ex5} does not imply (\tb{H}). The source of this difference is the different types of negation used. In the case of the implicative verb \emph{manage that} \citep{karttunen1971implicative}, some of its meaningful components are not within the scope of internal negation, e.g., the information that someone tried to do something. Thus, this information is actualized in the utterance~\ref{ex:ex4} in the form of information (\tb{H}). The external negation applied to the utterance~\ref{ex:ex5} makes it possible for all the meaning components of the verb to be within its range, so (\tb{H}) does not follow from~\ref{ex:ex5}.

\subsubsection{Presupposition}
We understand the term \emph{presupposition} as follows: If the utterance (\tb{T}) entails (\tb{H}) and (\tb{H}) is insensitive to internal negation, whether currently occurring or potential, then (\tb{H}) is the presupposition of the utterance (\tb{T}). We named all such information as presuppositions, regardless of their detailed nature. Thus, \tb{presuppositions can have both semantic and non-semantic grounds}. In the literature, one can find lists of expressions and constructions that are classically called \emph{presupposition triggers} \citep{levinson1983pragmatics}. Below is an example illustrating a presupposition based on a factive verb. 
\begin{example} \label{ex:ex8}\\
\noindent \tb{T}: [PL] \emph{Ona wie, że należy podać hasło.} \\
\noindent \tb{T}: [ENG] \emph{She knows that a password must be provided.} \\
>> \tb{H} [PL]: Należy podać hasło. \\
>> \tb{H} [ENG]: A password must be provided. 
\end{example}

Information (\tb{H}) in Example~\ref{ex:ex8} is the presupposition of the utterance  because this information is insensitive to internal negation. Presupposition (\tb{H}) is guaranteed by the factivity property of the verb \emph{wiedzieć, że / know that}.

The relation  between a semantic entailment and a semantic presupposition is shown in Example~\ref{ex:ex10}.\footnote{'$\models$' -- lexical entailment; '$\gg$' -- presupposition; '$/\gg$' -- no presupposition} \\
\begin{example} \label{ex:ex10} \\
\noindent \tb{T}: The driver managed to open the door before the car sank. \\
$\models\land/\gg$ (\tb{H}a) The driver managed to open the door before the car sank.\\
$\models\land\gg$ (\tb{H}b) The driver tried to open the door before the car sank.\\
$\models\land/\gg$ (\tb{H}c) The driver opened the door before the car sank.\\
$\models\land\gg$ (\tb{H}d) The car sank.
\end{example}

The utterance in Example~\ref{ex:ex10} semantically entails (\tb{H}a)-(\tb{H}d) because the definition of this type of entailment is fulfilled: if \tb{T} is true, then \tb{H} must be true. Apart from that, the information (\tb{H}b) and (\tb{H}d) are also presuppositions of the utterance. They meet the defining conditions of a presupposition: they are insensitive to -- in this case, potential -- internal negation. In other words, we treat presupposition as a certain subtype of entailment.

Analyzing the difference between semantic presupposition and non-semantic presupposition, consider the following utterances in Examples~\ref{ex:ex11} and \ref{ex:ex12}. Both utterances in Examples~\ref{ex:ex11} and \ref{ex:ex12} entail information (\tb{H}). Moreover, in both cases, (\tb{H}) is not within the scope of internal negation, so the information (\tb{H}) is their presupposition. However, \tb{the foundations of these presuppositions are radically different}. In Example~\ref{ex:ex11}, (\tb{H}) is guaranteed by the appropriate prosodic structure of the utterance, whereas in Example~\ref{ex:ex12}, the presupposition (\tb{H}) has a semantic grounding -- it is guaranteed because of the factive verb \emph{know that}. In other words, the entailment in Example~\ref{ex:ex11} is not lexical, and in Example~\ref{ex:ex12}, it is.
\begin{example} \label{ex:ex11} \\
\noindent \tb{T}: \emph{She was not told that he was already married.} \\
$\gg$ \tb{H}: He is already married. 
\end{example}
\begin{example} \label{ex:ex12} \\
\noindent \tb{T}: \emph{She didn't know that he was already married.} \\
$\gg$ \tb{H}: He is already married.
\end{example}

\subsubsection{Factivity}
It is worth noting the occurrence of the following four terms in the NLI literature: \emph{factivity}, \emph{event factuality}, \emph{veridicality}, \emph{speaker commitment}. These terms, unfortunately, are sometimes understood differently depending on the work in question. In the presented study, we use only the term \emph{factivity}, which is understood as an element of the meaning of particular lexical units. Such phenomena as speaker "degrees of certainty" are entirely outside the scope of the research presented here. We assume that presupposition founded on factivity takes place independently of communicative intentions; it may or may not occur: there are no states in between. For comparison, let's look at how the term "event factuality" is understood by the authors of the FactBank dataset:

\noindent \emph{"(…) we define event factuality as the level of information expressing the commitment of relevant sources towards the factual nature of events mentioned in discourse. Events are couched in terms of a veridicality axis that ranges from truly factual to counterfactual, passing through a spectrum of degrees of certainty \citep[p.231]{sauri2009factbank}."}

In another paper, the same pair of authors provide the following definition:

\noindent \emph{"Event factuality (or factivity) is understood here as the level of information expressing the factual nature of eventualities mentioned in text. That is, expressing whether they correspond to a fact in the world (…) \citep[p.263]{sauri2012you}."}   
It seems that the above two explanations of the term event factuality are significantly different. They are also composed of other terms that require a separate explanation, e.g. \emph{discourse}, \emph{veridicality}, \emph{spectrum of degrees of certainty}, \emph{level of information}.

Note that the second quoted fragment also defines factivity; the authors apparently put an equality mark between "event factuality" and "factivity" Reading the specifications of the FactBank corpus and the instructions for the annotators leads, in turn, to the conclusion that Saurí and Pustejovsky understand factivity as (a) a subset of factuality, (b) in a "classical way", as a property of certain lexical entities~\citep{sauri2009factbank}.

In the presented approach, factivity is understood precisely as a semantic feature of specific lexical units. In other words, it is an element of their meaning. According to the terminology used in the literature, we will say that factive verbs are \emph{presupposition triggers}. Using the category of semantic signature, it can be said that factive verbs belong to the category (+/+) \footnote{See Section~\ref{sec:related-datasets} for the explanation of the term \emph{signature}.}. Examples~\ref{ex:ex13} and \ref{ex:ex14} illustrate presuppositions based on the meaning of the factivity verb \emph{be aware that}.
\begin{example} \label{ex:ex13} \\
\noindent \tb{T}: \emph{She was not aware that she had made a fatal mistake.} \\
$\models$ \tb{H}: She made a fatal mistake.  
\end{example} 
\begin{example} \label{ex:ex14} \\
\noindent \tb{T}: \emph{She was aware she had made a fatal mistake.} \\
$\models$ \tb{H}: She made a fatal mistake.  
\end{example}

Information (\tb{H}) follow semantically from both Example~\ref{ex:ex13} and its modification as Example~\ref{ex:ex14}. This information is beyond the scope of internal negation and is, therefore, its presupposition. The foundation of this presupposition is the presupposition trigger in the form of the verb \emph{be aware that}. Neither the common knowledge of the speakers nor the prosodic properties of the utterances are irrelevant to the fact that (\tb{H}) in the above examples are presuppositions.

In summary, presuppositions can be either lexical or pragmatic in nature. What they have in common is that they are insensitive to internal negation. We treat presuppositions founded on factive verbs as lexical presuppositions. If information (\tb{H}) is a lexical presupposition of an utterance \emph{T}, then \emph{T} entails (\tb{H}). These relations are independent of the speaker's communicative intention; it means that the speaker may presuppose something unconsciously or against his own communicative intention.

\section{Related Datasets}
\label{sec:related-datasets}

The historical background of this paper is gathered from~\citep{cooper1996fracas,dagan2005pascal,dagan2013recognizing}. These works established a certain pattern of construction of linguistic material, consisting in pairs of sentences: \emph{thesis} and \emph{hypothesis} (\textbf{<T, H>}). In this work, the source of (\tb{T}) utterances is NKJP, and  \tb{H}) are complement clauses of (\tb{T}).
\begin{example} \label{ex:ex15} \\
\noindent \tb{T}: [PL] Myśleli, że zwierzęta gryzą się ze sobą. [NKJP] \\ 
\tb{H}: Zwierzęta gryzą się ze sobą. \\
\tb{T}: [ENG] They thought the animals were biting each other. \\
\tb{H}:	The animals were biting each other. 
\end{example}

We will now review some of the most recent and similar works. The first of these is \citep{ross-pavlick-2019-well}. The central term of this work is \emph{veridicality}, which is understood as follows: \emph{"A context is veridical when the propositions it contains are taken to be true, even if not explicitly asserted."}~\citep[p. 2230]{ross-pavlick-2019-well}. As can be seen, the quoted definition also includes situations in which the entailement is guaranteed by factive verbs. The authors pose the following research question: 'whether neural models of natural language learn to make inferences about veridicality consistent with those made by humans? This is a very different question from the one posed in this paper. Ross and Pavlick used Likert scales to conduct annotations employing unqualified annotators~\citep{ross-pavlick-2019-well}. They then checked the extent to which the models' predictions coincide with the human annotations obtained.  Unlike these authors, we do not use any scales in annotation, and the object of the models we train is to predict real semantic relations, not those that occur as judged by humans. 

Let's also note that this is an example of work that uses semantic signatures. They distinguish between eight pairs of semantic signatures, e.g., (+/+) (\emph{realize that}), (+/-) (\emph{manage to}), (-/+) (\emph{forget to}). A similar approach is in \citep{rudinger-etal-2018-neural-models}, i.e., factive verbs are one of several types of expressions that are of interest. In contrast to these works, we have distinguished \textbf{only two classes of verbs: factive (+/+) and non-factive $\neg$(+/+).} Thus we included verbs that belong to the group "$\neg$(+/+)", i.e. verb classes such as (+/-), (-/-), (-/+) etc. 
Due to the fact that we operate with the concepts of factive/non-factive verbs, we do not use the notion of semantic signatures in this paper. We are aware that in similar papers the number of verb distinctions is sometimes significantly higher. The decision to use only  a binary distinction (factive vs. non-factive) is dictated by several interrelated considerations. First, there are no lists of Polish verbs with signatures assigned by linguists. Secondly, the preparation of such a list of even several dozen verbs is a highly specialized task. It may be added that there are still disputes in the literature about the status of some high frequency verbs, e.g. \emph{regret that}. Third, we are interested in the real features of lexical units, and not in the 'textual' ones, i.e. those developed by non-specialist annotators, using the committee method. The development of implication signatures by unqualified annotators would be pointless with regard to the research questions posed. The type of linguistics used in this work is formal linguistics, which investigates the real features of a language, unlike survey linguistics, which collects the intuitions of speakers of an ethnic language. \footnote{See \citep{ipeirotis2010quality} and \citep{hsueh2009data} on the problems of low quality annotation with Amazon Turk and how to solve them.} The last reason is that the factivity/non-factivity split, given the frequency of occurrence of these relations, is most important for entailment and neutral.

Another important paper is \citep{jiang-de-marneffe-2019-evaluating}, in which the authors take a closer look at the CommitmentBank dataset. Also, in the dataset, the annotation process used a Likert scale. The paper concludes that BERT systematically commits specific patterns of errors; it does not handle inferences described as pragmatic, e.g.: \\
\tb{T}: \emph{Those people... Not a one of them realized I was not human. They looked at me and they pretended I’m someone called Franz Kafka. Maybe they really thought I was Franz Kafka.} \\
\tb{H}: \emph{he was Franz Kafka} \\

Other new work worth paying attention to is \citep{parrish2021nope}. It presents the Naturally-Occurring Presuppositions in English (NOPE) Corpus that considers a diverse set of presupposition triggers (as many as 10 types). It is worth noting that this set does not include factivity. The authors argue that there is no clear distinction between factive and non-factive verbs.They also state that even verbs such as know, which "are commonly regarded as factive," do not always guarantee that the complement is true.  We strongly disagree with the claim that there is no clear distinction between factive and non-factive verbs: this is a central assumption of our work. We do, however, of course agree that in certain contexts factive verbs do not guarantee the truthfulness of sentence complements: however, we treat this as a research challenge; at the current stage of work, we estimated the scale of this phenomenon in communication conducted in Polish.

It is also worth noting completely new work on the topic of interest, namely \citep{jiang2021he}, \citep{tarunesh2021trusting}, \citep{tarunesh2021lonli}, \citep{yanaka2021exploring} . These works, like the earlier ones, point to the limited possibilities of recognizing phenomena such as facticity.

\section{Language Material \& Our Dataset} 
\label{sec:dataset}

Our \dataset   dataset focuses on a specific linguistic phenomenon: the opposition of factivity vs. nonfactivity and the relation of these categories to semantic features such as entailment, contradiction and neutrality. We conclude that the specified datasets allow for a better specialization of ML models to narrow their scope of features to generalize (see \citep{poliak-2020-survey}). The three most important features of our dataset are as follows:
\begin{itemize}
\item it does not contain any prepared utterances, only authentic examples from the national language corpus, 
\item it is not balanced, i.e., some features are represented more frequently than others; it reflects authentic communication in Polish, 
\item each pair <T, H> is assigned a number of linguistic features, e.g. the main verb, its grammatical tense, the presence of internal negation, the type of utterance, etc. In this way, it allows us to compare different models -- embedding-based or feature-based. 
\end{itemize}

\subsection{Input Material Sources \& Extraction}
The material basis of our dataset is the National Corpus of Polish Language (NKJP)~\citep{Przepiorkowski:12}. We used a subset of NKJP in the form of the PKK Polish Coreference Corpus (PKK)~\citep{ogrodniczuk2014coreference}), which contains randomly selected fragments from NKJP and constitutes its representative sample. We did not add any prepared utterances -- our dataset consists only of original utterances found in the PKK. Moreover, the selected utterances have not been modified in any way -- we did not correct typos, syntactic errors, or other difficulties and issues. Thus, the language content remained entirely natural, not improved artificially.

We automatically annotated with Discann\footnote{http://zil.ipipan.waw.pl/Discann} all occurrences of the phrase \emph{że} (\emph{that} | \emph{to}) as in Example~\ref{ex:ex16}.
\begin{example} \label{ex:ex16} \\
\noindent \tb{T}: [PL] \emph{Przez lornetkę obserwuję, że zalane zostały żerowiska bobrów.}\\
\tb{T}: [ENG] \emph{I can see through binoculars that the beaver feeding grounds have been flooded.}
\end{example}

From more than 3,500 utterances, we left only those that satisfied the following pattern: '{[verb] [że] [complement clause]}.' It required a manual review of the entire dataset. Thus, we obtained \textbf{2,320} utterances that constitute the empirical basis of our dataset.

\subsection{Dataset Content}
Finally, the dataset consists of 2,320 real utterances from which 2,432 \textbf{<T, H>} pairs were formed. Each of these pairs was assigned one of three relations: entailment, contradiction, and neutral. In addition, each utterance \textbf{<T>} was assigned several linguistic features. The occurrence of features is not balanced, i.e. \textit{entailment} class states $33.88\%$ of the dataset, \textit{contradiction} -- $4.40\%$, \textit{neutral} -- $61.72\%$. Thus, in this shape, the dataset constitutes a representative sample of communication in Polish. As can be seen, e.g., utterances with negation in the studied syntactic construction appear relatively rarely (less than 5\%). Table~\ref{tab:dataset-statistics} shows the detailed distribution of features in the set. 
\begin{table*}[!htbp]
\caption{Distributions of features in \dataset dataset in Polish version\label{tab:dataset-statistics}}
\centering
\begin{tabular}{p{4.5cm}p{8cm}}
\toprule
Features & Distribution \\
\midrule
\textbf{target / GOLD -- logic relations} &  \textit{entailment} 33.88\%; \textit{contradiction} 4.40\%; \textit{neutral} 61.72\% \\
Verb type (factivity) &  factive 24.96\%; non-factive 75.04\% \\
Grammatical verb tense & past 36.18\%; present 52.22\%; future 3.08\%; none 8.51\%  \\
Utterance type &  indicative 90.17\%; performative 2.43\%; rule 2.22\%; interrogative 1.97\%; imperative 1.93\%; counterfactual 0.66\%; conditional 0.62\% \\
Verb semantic class & epistemic 51.85\%; speech 38.03\%; perceptual 1.81\%; emotive 1.40\%; other 6.91\%; \\
Occurrence of internal negation  &  occurs in 4.93\%; does not occur in 95.07\% \\
Tense of complement clause  & past 23.23\%; present 49.01\%; future 14.80\%; other 12.95\%  \\
\bottomrule
\end{tabular}
\end{table*}

\begin{table*}[!htbp]
\caption{Contingency table consisting of the frequency distribution of two variables.\label{tab:factive_vs_classes}}
\centering
\begin{tabular}{p{2.5cm}p{2.5cm}p{2.5cm}}
\toprule
 & \textbf{Factive} & \textbf{Non-factive} \\ \midrule
\textbf{C} -- contradiction & 0 & 107 \\ 
\textbf{E} -- entailment & 593 & 231 \\ 
\textbf{N} -- neutral  & 14  & 1487\\
\bottomrule
\textbf{Total} & 607 & 1825 \\
\bottomrule
\end{tabular}
\end{table*}

\subsection{Linguistic Features}

\subsubsection{Utterances \& Information -- <T, H> Pairs}
From 2,320 utterances, we created 2,432 \textbf{<T, H>} pairs (309 of unique main verbs). In some utterances, the verb introduced more than one complement -- in each such situation, we created a separate \textbf{<T, H>}. For each sentence, we extracted a complement clause manually. Our manual work included, e.g., removing fragments that were not in the range of the verb -- see Example~\ref{ex:ex17}.
\begin{example} \label{ex:ex17} \\
\noindent \tb{T}: He said I am -- I guess -- beautiful.\\
\tb{H}: I am \sout{- I guess -}  beautiful.
\end{example}

The pairs we have created \textbf{<T, H>} are the core of our dataset. We assigned specific properties (i.e., linguistic features) to each pair. In the following, we presented these linguistic features with their brief characteristics.

\subsubsection{Entailment, Contradiction and Neutral (ECN)}
The process of annotating (ECN) relations was performed by an expert linguist experienced in natural language semantics and then checked by another expert in formal semantics.

\subsubsection{Verb}
In each utterance, the experienced linguist manually identified the verb that introduced the H sentence.

Despite appearances, this was not a trivial task. Often identifying a given lexical unit required deep thought and verification of specific delimitation hypotheses. Among other things, in order to avoid problems of polysemy, we assumed that \textbf{one meaning} can be assigned to a given verb, e.g., we distinguish between \emph{czuć, że} / \emph{feel that}, which is epistemic, and \emph{czuć, że} / \emph{feel that}, which is purely perceptual (see Examples~\ref{ex:ex18} and \ref{ex:ex19}).

\begin{example} \label{ex:ex18} (Epistemic) \\
\noindent \tb{T}: \emph{He felt that he would never see it again}.
\end{example}
\begin{example} \label{ex:ex19} (Perceptual) \\
\noindent \tb{T}: \emph{He felt that he was walking in his arms.}
\end{example}

We identified a total of 309 verbs.\footnote{We treated the aspect pairs as two verbs. The Polish language, in a nutshell, has only two aspects: perfect and imperfect.} 

\subsubsection{Verb Type}
We assigned  one of two values: \emph{factive} / \emph{non-factive} to all verbs. From the linguistic side, it was the most difficult part. This task was done by a linguist writing his Ph.D. thesis on the factivity phenomenon. The list was checked with the thesis supervisor, and in most cases, these people agreed with each other, but not in all cases. Finally, 81 verbs were marked as factive and 230 as non-factive.

\subsubsection{Internal Negation}
For each utterance, we marked whether it contains an internal negation. About 95\% utterances did not contain explicit negation words, and almost 5\% sentences did.

\subsubsection{Verb Semantic Class}
We have distinguished four semantic classes of verbs: epistemic (\emph{myśleć, że} / \emph{think that}), speech (\emph{dodać, że} / \emph{add that})), emotive \emph{żałować, że} / \emph{regret that}) and perceptual \emph{dostrzegać, że} / \emph{perceive that}). Most verbs were hybrid objects, e.g. epistemic-speech. The class name was given due to the given dominant semantic component. If the verb did not fit into any of the above classes, the value \emph{other} was given.

\subsubsection{Grammatical Tense}
In each utterance, we marked the grammatical tense of the verb and the complement \textbf{H}.

\subsubsection{Utterance Type}
We labeled the type of utterance as: indicative, imperative, counterfactual, performative, interrogative, or conditional.

All \textbf{T} utterances have been translated into English, see Appendix~\ref{sec:sup:translation}.

\subsection{Annotation}
Among linguistic features assigned to pairs \textbf{<T, H>} the most difficult and essential to identify were factivity/non-factivity and logic relations ECN. Whether a verb is factive was determined by two linguists who are professionally involved in natural language semantics. They achieved more than 90\% agreement, with most doubts arising when analyzing verbs of speaking, e.g., \emph{wytłumaczyć, że} / \emph{explain that}. The final decisions on identifying which verb is factive were made by a linguist writing a PhD on the topic of factivity in contemporary Polish and it was checked by his supervisor -- a professor of formal linguistics.

Semantic relations ECN were established in two stages. The first stage was annotated by two linguists, including one who academically deals with the issue of verb semantics. They achieved 70\% agreement in the annotation. Significant discrepancies can be observed for relations of contradiction, as opposed to entailment. Then, those debatable pairs were discussed with a third linguist, a professor specializing in natural language semantics. The result of these consultations was the final annotation - the GOLD standard.

We checked how the GOLD standard created in this way would differ from the annotations of non-professional linguists -- a group of annotators who are not involved professionally in formal linguistics but have a high linguistic competence. The criteria for selecting annotators were the level and type of education and a pre-test. Thus the four selected annotators were: (1) cognitive science student (3rd year), (2) and (3) master's degree in Polish Studies, master's degree in Polish Studies, (4) linguistics Ph.D.

Each annotator was given the same set of <T,H> pairs from the set (20\% of the total set). The task of each annotator was to note the relation between \tb{T} and \tb{H}. There were four labels to choose from: \textit{entailment}, \textit{contradiction}, \textit{neutral} and '\textit{?}'.\footnote{However, the utterances annotated as "?" in the GOLD label, were not taken for training and testing ML benchmarks.} 
The annotation instructions included simplified definitions of key labels -- as we presented in Section~\ref{sec:terms}.

Annotators were asked to choose '\textit{?}', if: (1) they could not indicate what the relation was, or (2) they thought the sentence was meaningless, or (3) they encountered another problem that made it impossible for them to choose any of the other three labels. Especially important, from our point of view, is the situation (1). The idea was to reserve it for $<$\tb{T}, \tb{H}$>$ pairs whose semantic relation is dependent on prosodic features (like accent, which determines focus and topic~\citep{Partee_1991}.  Let's look at an example ~\ref{ex:ex20}:
\begin{example} \label{ex:ex20} \\
\noindent \tb{T}: \emph{Let's not say [that] these projects are supposed to end in a constitutional change.} \\
\tb{H}: These projects are supposed to end in a constitutional change. 
\end{example}

There are two possible situations in Example~\ref{ex:ex20}: (a) the sender wants to hide the information from \tb{H} (label: \textit{entailment}) and (b) the sender does not want to say \tb{H} because, e.g., he wants to make sure that this is true first (label: \textit{neutral}). 

Inter-annotator agreement with the dataset gold standard was in the range of 61\% -- 65\%, excluding the worst annotator whose Kappa was below 52\% with all other annotators.\footnote{A few annotation examples are given in our supplement~\ref{sec:sup:examples}}
Table~\ref{tab:agreement} summarizes the inter-annotator agreement among four non-expert linguists and one of the experts preparing the dataset. The conclusions of the annotation performed and described above are provided in Section~\ref{subsection:annotation_task}.

\begin{table}[!htb]
\caption{Inter-annotator agreement given by Cohen's Kappa (alpha=0.05). Note: Ex -- an expert who made the gold standard, A1-A4 -- non-expert linguists. \label{tab:agreement}}
\centering
\begin{tabular}{llllll}
\toprule
 & Ex      & A1      & A2      & A3      & A4      \\
\midrule
Ex & 1.00 & 0.65 & 0.60  & 0.38  & 0.61  \\
A1 & 0.65  & 1.00 & 0.59  & 0.29  & 0.51  \\
A2 & 0.60  & 0.59  & 1.00 & 0.47  & 0.70  \\
A3 & 0.38  & 0.29  & 0.47  & 1.00 & 0.52  \\
A4 & 0.61  & 0.51  & 0.70  & 0.52 & 1.00 \\
\bottomrule
\end{tabular}
\end{table}

\section{Machine Learning Modelling --  Experiments and Results}
\label{sec:ML-modeling}

The models we built aim to simulate human cognitive abilities. The models trained on our dataset were expected to reflect high competence -- comparable to that of human experts -- in recognizing the relations of entailment, contradiction, and neutral (ECN) between an utterance (\emph{T}) and its complement (\emph{H}). We trained five kinds of models:
\begin{enumerate}
    \item[(1)] Random Forest with an input of the prepared linguistic features, 
    \item[(2)] fine-tuned HerBERT-based models for only main verbs in sentences as inputs, 
    \item[(3)] model (2) with input extended with linguistic features, 
    \item[(4)] fine-tuned HerBERT-based model for the whole input utterance (\emph{T}), 
    \item[(5)] model (3) with input extended with linguistic features. 
\end{enumerate}
We employed HerBERT~\citep{rybak-etal-2020-klej} models instead of BERT, because they are trained explicitly for Polish and achieved better results in comparison to Polish RoBERTa~\citep{rybakKLEJ}. Python code and \dataset  dataset can be found in our GitHub repository~\footnote{\small{\textit{https://github.com/grant-TraDA/factivity-classification}}}.

Each model was trained using 10-fold cross validation in order to avoid selection bias. Table~\ref{tab:overall-results} shows the models' results achieved on the first seen data (unknown data for a model). The values in the table represent mean and standard deviation of metrics, respectively. 
F1 score in binary setting is harmonic mean between model precision and recall. In multiclass situation it is calculated per class and overall metric is average score. Here F1 score was calculated as weighted F1, due to large imbalance between classes.

\begin{table*}[!htbp]
\caption{Classification results of entailment (E), contradiction (C) and neutral (N). Linguistic features comprise: verb, grammatical tense of verb, occurrence of internal negation, grammatical tense of complement clause, utterance type, verb semantic class, verb type  (factive/non-factive). F1  score depicts weighted F1 score. \label{tab:overall-results}}
\centering
\begin{tabular}{p{1.3cm}p{2.7cm}p{1.8cm}p{1.9cm}p{1.8cm}p{1.8cm}|p{1.8cm}}
\toprule
\multirow{2}{*}{Model} & \multirow{2}{*}{Input} & \multicolumn{4}{p{4cm}|}{F1 score [\%]} & \multirow{2}{*}{Accuracy [\%]} \\ 
& & All & C & E & N & \\ \midrule 
Random Forest & Linguistic features  & $90.56 \pm 2.11$ & $39.89 \pm 23.30$ & $91.96 \pm 2.50$ & $93.35 \pm 1.32$& $91.32 \pm 1.84$ \\ \midrule
HerBERT & Verb embedding & $89.45 \pm 1.00$ & $39.21 \pm 17.57$ &	$90.50 \pm 1.93$ &	$92.42 \pm 0.75$ & $90.21 \pm 1.06$ \\ \midrule
HerBERT & Verb embedding + linguistic features & $90.52 \pm 2.37$ &	$33.00 \pm 31.97$	& $92.45 \pm 1.88$ &	$93.53 \pm 1.36$ & $91.53 \pm 1.89$ \\ \midrule
 HerBERT & Sentence embedding & $88.51 \pm 1.06$ & $48.33 \pm 11.93$ & $88.34 \pm 1.61$ &	$91.46 \pm 0.08$ & $88.57 \pm 1.11$ \\ \midrule
HerBERT & Sentence embedding + linguistic features & $89.92 \pm 1.23$ & $26.25 \pm 17.00$ & $92.27 \pm 1.84$ &	$93.15 \pm 0.71$ & $90.95 \pm 1.15$ \\ 
\bottomrule
\end{tabular}
\end{table*}

\begin{figure}[!htb]
 \includegraphics[width=13cm]{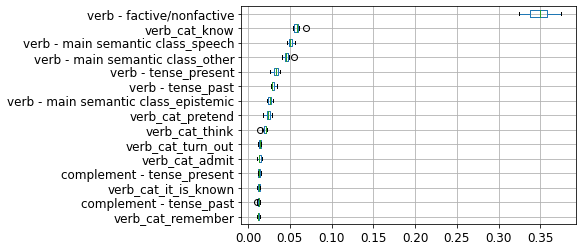}
 \vspace{25pt}
\caption{Impurity-based feature importance of  feature-based Random Forest. The chart shows the English equivalents of Polish verbs:  \emph{know}/\emph{wiedzieć}; \emph{pretend}/\emph{udawać}; \emph{think}/\emph{myśleć}; \emph{turn out}/\emph{okazać się}; \emph{admit}/\emph{przyznać}; \emph{it is known}/\emph{wiadomo}; \emph{remember}/\emph{pamiętać}.  \label{fig:feature-importance}}
 \end{figure}

The parameters of models and their training process are gathered in  Table~\ref{tab:model-params}.
The precise results of Random Forest for different feature sets and the feature importance plots are given in Table~\ref{tab:feature-models} and in Figure~\ref{fig:feature-importance}.  Table~\ref{tab:characteristic-classes} summarises the results of our models for the most characteristic sub-classes in our dataset: entailment and factive verbs, neutral and non-factive verbs, and the other cases.

\begin{table}[!htbp]
\caption{Model and training parameters.\label{tab:model-params}}
\centering
\begin{tabular}{p{0.5cm}p{4cm}p{7.5cm}}
\toprule
& Model & Parameters \\ \midrule
1 & Random Forest & sklearn implementation with 100 trees (n\_features=100, max\_depth=20, random\_state=123, class\_weight=\{'C': 2, 'E': 1, 'N': 1\} and default other parameters)  \\ 
2 & HerBERT (verb embedding) & 32 - batch size, 10 - epochs, 1e-5 - learning rate (Pytorch implementation of Adam) \\ 
3 & HerBERT+linguistic features (verb embedding) & Predictions of model (2) combined with linguistic features preprocessed with one hot encoding, sklearn implementation of Multi-layer Perceptron  \\
4 & HerBERT (sentence embedding) & 32 - batch size, 13 - epochs, 1e-5 - learning rate (Pytorch implementation of Adam) \\ 
5 & HerBERT+linguistic features (sentence embedding) & Predictions of model (4) combined with linguistic features preprocessed with one hot encoding, sklearn implementation of Multi-layer Perceptron \\
\bottomrule
\end{tabular}
\end{table}

\begin{table*}[!htbp]
\caption{Features in Classification of Entailment, Contradiction and Neutral. Random Forest results with inputs of different sets of features.\label{tab:feature-models}}
\centering
\begin{tabular}{p{9cm}p{2cm}p{2cm}}
\toprule
Features &  Accuracy [\%] & Weighted F1 [\%] \\ \midrule
verb -- factive/non-factive & $85.53 \pm 1.78$ & $83.24 \pm 1.92$ \\ \midrule
verb & $83.23 \pm 2.29$ & $81.29 \pm 2.75$ \\ \midrule
verb ,
    tense of verb,
    occurrence of negation,
    tense of complement clause,
    type of sentence & $83.10 \pm 1.93$ &  $81.67 \pm 2.38$\\ \midrule
verb ,
    tense of verb,
    occurrence of negation,
    tense of complement clause,
    type of sentence, \textbf{semantic class of verb} & $86.76 \pm 1.44$ & $85.83 \pm 1.89 $ \\ \midrule
verb ,
    tense of verb,
    occurrence of negation,
    tense of complement clause,
    type of sentence, \textbf{semantic class of verb}, \textbf{verb - factive/non-factive} & $91.32 \pm 1.84$ & $90.56 \pm 2.11$ \\
\bottomrule
\end{tabular}
\end{table*}

\begin{table*}[!htbp]
\caption{Results in the most characteristic subsets in our test dataset: entailment and factive, neutral and non-factive, and the other cases. \label{tab:characteristic-classes}}
\centering
\begin{tabular}{p{2cm}p{2.1cm}p{2cm}p{2cm}p{2cm}p{2cm}}
\toprule
Model & Random Forest model accuracy [\%] & HerBERT-based (sentence embedding) model accuracy [\%]  & HerBERT-based (sentence embedding + linguistic features) model accuracy [\%] & HerBERT-based (verb embedding) model accuracy [\%] & HerBERT-based (verb embedding + linguistic features) model accuracy [\%] \\ \midrule
Entailment and factive verbs &	$99.81 \pm 0.56$
&  $93.00 \pm 2.76$ & $100.00 \pm 0.0$& $97.58 \pm 2.54$ & $100.00 \pm 0.0$ \\
Neutral and non-factive &	$98.19 \pm 1.12$ & $93.01 \pm 1.83$ & $96.77 \pm 1.23$ & $95.75 \pm 1.95$ & $97.04 \pm 1.28$\\
Others & $47.76 \pm 10.18 $	&
$62.87 \pm 6.04 $& $50.81 \pm 7.32$ & $53.81 \pm 9.71$ &  $52.15 \pm 9.71$\\
\bottomrule
\end{tabular}
\end{table*}

\section{Results Analysis \& Discussion}
\label{sec:Results Analysis/Discussion}
\subsection{Issues in Dataset Preparation}
We gathered dataset \dataset that is representative with regard to particular syntactic pattern '[verb][że (eng: that/to)][complement clause]' and factivity and non-factivity characteristic of the verb (in the main clause). The representation dataset is derived from NKJP -- Polish national corpus, which itself is representative for Polish contemporary utterances. Thus, based on this material -- our dataset -- we can answer our first research question RQ1, which, recall, was as follows:
\begin{enumerate}
    \item[RQ1:] \emph{How relevant is the opposition factivity / non-factivity for the prediction of ECN relations?}
\end{enumerate}
\textbf{Firstly}, the distribution of features in the dataset indicates that, in the vast majority of cases, factive verbs go with an entailment relation (24.4\% of our dataset), and non-factive verbs with a neutral relation (61.1\%) -- see Table~\ref{tab:factive_vs_classes}. Other utterances, i.e., the pairs \textbf{<T, H>}, in which, for example, despite a factive verb, there is neutral, or despite a non-factive verb, there is entailment, constitute a narrow subset of the dataset (14.5\% -- 352 utterances in the dataset). Table~\ref{tab:hard-examples} contains examples of such pairs. These kinds of \textbf{<T, H>} pairs pose the biggest problem for humans and models -- the best model accuracy of 62.87\% (see Table~\ref{tab:characteristic-classes}). Let's recap - in 85,5\% of the pairs of the whole dataset, entailment co-occurs with a factive verb or the neutrality co-occurs with a non-factive verb. 

\textbf{Second}, if the verb was factive, then the entailment relation occurred in 97,70\%. And if the verb was nonfactive, the neutrality occurred in 81,50\%. It means that pairs of features <factivity, entailment> and <non-factivity, neutrality> very often co-occur with each other, especially the first pair. This means that such phenomena as cancellation and suspension of presuppositions \footnote{See, for example, the paper ~\citep{abrusan2016presupposition}, which discusses the phenomena behind these terms.} are marginal in our dataset.

\begin{figure}
\centering

\begin{minipage}{.5\textwidth}
  \centering
  \includegraphics[width=1\linewidth]{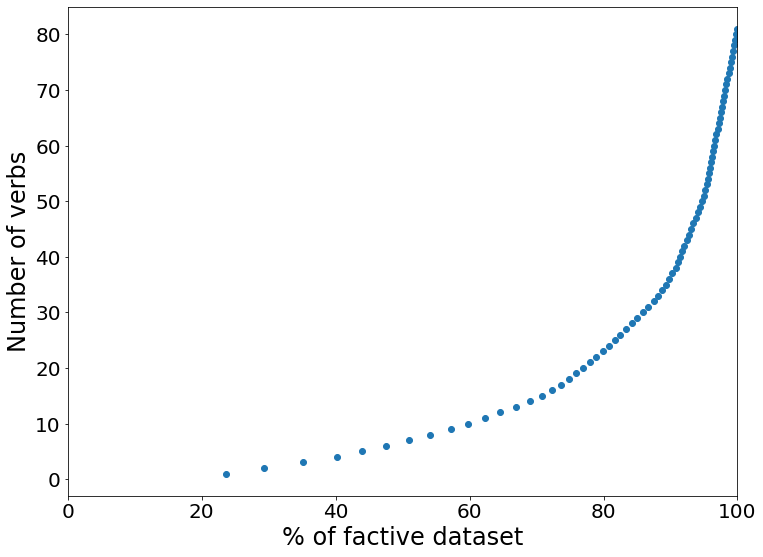}
  \label{fig:sub1}
\end{minipage}%
\begin{minipage}{.5\textwidth}
  \centering
  \includegraphics[width=1\linewidth]{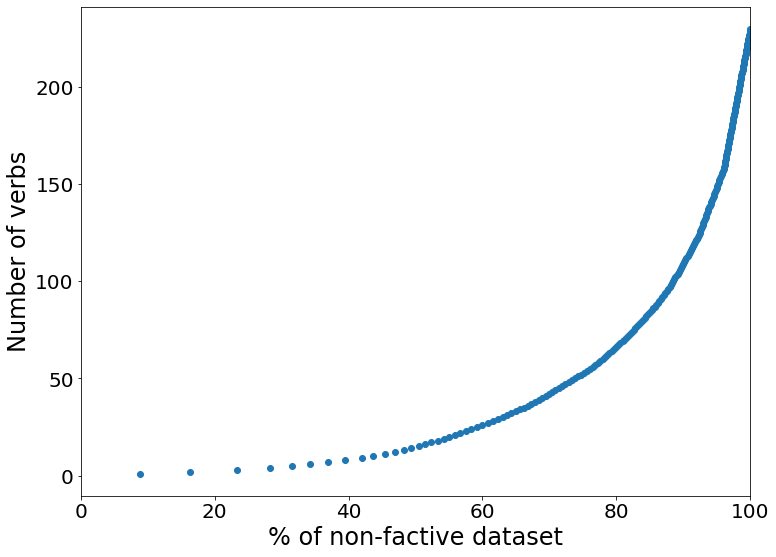}
  \label{fig:sub2}
\end{minipage}
\vspace{5pt}
\caption{Relationship between the number of the most frequent verbs and the coverage of dataset. Left: The analysis of factive subsample. Right: The analysis of non-factive subsample.}
\label{fig:factivity-test}
\end{figure}

\textbf{Thirdly}, according to our dataset, only 10 factive verbs with the highest frequency account for the occurrence of 60\% of all occurrences of such expressions, and 10 nonfactive verbs with the highest frequency account for nearly 45\% of all occurrences of nonfactive verbs (see Figure~\ref{fig:factivity-test}).

In view of the above, it can be concluded that \textbf{the opposition factivity / non-factivity for the prediction of ECN relations is relevant in a fundamental way}. In other words in the syntactic pattern under analysis, verbs with the signature (+|+) (factive) and the signature "$\neg$(+/+)" (non-factive) are most important in tasks that predict ECN relations between the whole utterance (\tb{T}) and its complement clause (\tb{H}). 

It is also worth noting that in the task of predicting ECN relations in the "V{that|to}p" stucture, we do not need large lists of verbs with their implicature signatures to identify these relations reasonably efficiently. Given the problem of translation of utterances from one language to another, it is therefore sensible to create a multilingual list of verbs with the highest frequency of occurrence. We realize that the frequency of occurrence of certain Polish lexical units may sometimes differ significantly from that of their equivalents in other languages. However, there are reasons to believe that these differences are not significant \footnote{Compare, for example, the verbs \emph{wiedzieć} and \emph{powiedzieć} with their English counterparts \citep{davies2010word} and many other such verbs realizing the "V{that|to}p" structure.}. A bigger issue than frequency may be a situation where a factive verb in language X is non-factive in language Y and vice versa.Table ~\ref{tab:top-verbs} contains lists the factive and non-factive verbs with the highest frequency in our dataset. We leave it to native speakers of English to judge whether the given English verbs are factive/non-factive.
\begin{table*}[!htbp]
\caption{Top 10 verbs broken down into factive / non-factive subgroups. \label{tab:top-verbs}}
\centering
\begin{tabular}{p{6cm}p{6cm}}
\toprule
\textbf{Factive} & \textbf{Non-factive} \\ 
\midrule
wiedzieć; know & mówić; say\\
pamiętać; remember & myśleć, ; think\\
wiadomo [komuś]; it is known & powiedzieć; tell\\
przyznać; admit/acknowledge & uważać; believe\\ 
widzieć; see [epistemic] & okazać się; turn out\\
cieszyć się; glad & mieć nadzieję ; hope\\ 
przypomnieć [komuś]; remind [someone] & twierdzić; assert\\
dowiedzieć się; find out/learn & wydaje się [komuś]; it appears [to someone]\\
zrozumieć; understand & stwierdzić; state\\
przyznawać; admit/acknowledge & wynikać; imply/follow
\\
\bottomrule
\end{tabular}
\end{table*}

At this point it is worth asking the following question: do the results obtained on the Polish material have any bearing on communication in other ethnic languages? We think it is quite possible. Firstly, the way of life and, consequently, the communication situations of the speakers of Polish do not differ from the communication situations of the speakers of English, Spanish, German or French. Secondly, we see no evidence in favor of a negative answer. It is clear, however, that the answer to this question requires research analogous to ours in other languages.
\begin{table*}[!htb]
\caption{Hard utterances in our dataset.\label{tab:hard-examples}}
\centering
\begin{tabular}{p{12.5cm}}
\toprule
\tb{T}: [ENG] \emph{How do you know he bought, Gosia?}

~~~~[PL] \emph{Skąd wiesz, że kupił, Gosia?}
\\
\tb{H}: [ENG] \emph{He bought.} 

~~~~[PL] \emph{Kupił.}\\
 GOLD -- Neutral, verb -- Factive \\
\midrule

\tb{T}: [ENG] \emph{I read that Gabrysia was crying when she discovered that her daughter Tygrysek had lied, and this is such a moral compass for me.} 

~~~~[PL] \emph{Czytam, że Gabrysia płakała, kiedy odkryła, że jej córka Tygrysek skłamała, i jest to dla mnie taki moralny azymut.}
\\
\tb{H}: [ENG] \emph{Her daughter Tygrysek had lied} 

~~~~[PL] \emph{Jej córka Tygrysek skłamała}\\
 GOLD -- Neutral, verb -- Factive \\
\midrule
\tb{T}: [ENG] \emph{Ernest and Agnieszka didn't plan that they would have a big female family.}

~~~~[PL] \emph{Ernest i Agnieszka nie planowali, że będą mieli wielką, babską rodzinę.}
\\
\tb{H}: [ENG] \emph{Ernest and Agnieszka have a big female family.} 

~~~~[PL] \emph{Ernest i Agnieszka mają wielką, babską rodzinę.}\\
GOLD -- Entailment , verb -- Non-factive \\
\bottomrule
\end{tabular}
\end{table*}

\subsection{Annotation Task}\label{subsection:annotation_task}

Inter-annotator agreement of non-expert annotators with the linguists'  preparing the dataset gold standard (Kappa of 61\% - 65\%) indicates that the task is very specialized. 
We did not find patterns of errors made by the annotators. If the goal of human annotation is to identify the real relationships between two fragments, then such annotation requires specialized knowledge and a sufficiently long time to perform such a task.

Note that \cite{jeretic2020natural}, as part of their verification of annotation in the MultiNLI corpus~\cite{williams-etal-2018-broad}, randomly selected 200 utterances from this corpus and presented them for evaluation to three expert annotators with several years of experience in formal semantics and pragmatics. The agreement among these experts was low. This provides the authors with an indication that MultiNLI contains a few "paradigmatic" examples of implicatures and presuppositions\footnote{See \cite{levinson2001pragmatics} for typical examples of presuppositions and implicatures.}. 

Notice that the low agreement of annotators may also be the result of differences in their beliefs of theoretical nature and their research specialization. In our opinion, the analysis of the issue of human annotation process in such a task as detecting relations of entailment, contradiction and neutral in principle deserves a separate study.

\subsection{ML Results Analysis}

Our ML experiment can help answer our second research question, RQ2, about assessing ML models in our task -- recognition of ECN relations. 
The conclusion derived from the dataset can be that classes Entailment and Neutral are most common in language and are the most important for models to deal with. The overall results indeed show very high performance of models in these classes: Entailment -- 88\% to 92\% in accuracy, and Neutral -- 91\% to 93.5\% in accuracy (see Table~~\ref{tab:overall-results}).
More precisely, the models achieved very high results (93\% up to 100\% in accuracy) for the sentence pairs containing lexical relations -- subsets entailment and factive, neutral and non-factive in Table~\ref{tab:characteristic-classes} -- but gained very low metrics (47 \% up to 62.9 \%) on those with a pragmatic foundation, which are drastically more difficult -- subset "Other" in Table~\ref{tab:characteristic-classes}.

Additionally, the overall ML modelling results show that HerBERT sentence embedding-based models are at a much higher level than non-expert linguists. Nevertheless, they did not achieve the results of professional linguists (by whom our GOLD standard was annotated). Feature-based models achieve slightly better results (mean accuracy across folds of 91.32\%), although not for the contradiction relation (mean accuracy of 39.89\%). However, the weak result for this relation is due to a small representation in the dataset (only 4.4\% cases, see Table~\ref{tab:dataset-statistics} and~\ref{tab:factive_vs_classes}). Regardless, according to the data obtained, contradiction in communication conducted in Polish occurs very rarely.
Moreover, the variance of the ML results is between 0.08 up to 2.5 \% across different folds in cross-validation for the overall results and the easier classes (E,N). However, the variance for C class (contradiction) is very high -- from 11.93 even up to 31.97 \%. Once again, this occurs because the phenomenon appears very rarely and in the dataset we have only a few cases (i.e. 107 cases).

Note that the performance of models trained on such a training set achieves significantly higher results than those obtained by annotators in the test annotation performed.This means that under certain conditions the trained models make human annotation work almost completely redundant.

Further, models with verb embedding vs the entire sentence representations are better. However, they require manual extraction of the main verb in the utterance because sometimes it is not apparent. In the following are examples of such a difficult extraction.

\begin{example} \label{ex:ex21} (Neutral)

\noindent \tb{T}: [PL] \emph{Czuł, że inni zbliżali się do niego, ale nie był tego pewien.} \\
\tb{T}: [ENG] \emph{He felt others getting closer to him, but he wasn't sure.} \\
\end{example}

\begin{example} \label{ex:ex22} (Entailment)

\noindent \tb{T}: [ENG] \emph{He felt that the typist wasn't taking her eyes off him...} \\
 \tb{T}: [PL] \emph{Czuł, że maszynistka nie spuszcza zeń oczu.} \\
\end{example}

Consideration of the difference between the main verbs in the above examples requires attention to suprasegmental differences. In Example~\ref{ex:ex21}, the non-factive verb \emph{czuć, że} is used. In contrast, in example B, a different verb is used, namely the factive \emph{CZUĆ, że}, which necessarily takes on the main sentence stress (see ~\cite{danielewiczowa2002wiedza}). Note that from the two verbs above, we should also distinguish the factive perceptual verb \emph{czuć, że}.

\begin{example} \label{ex:ex23} (Neutral)

\noindent \tb{T}: [PL] \emph{Lawirował na granicy prawdy, lecz przez cały czas czułem, że kłamie.} 
\end{example}

\begin{example} \label{ex:ex24} (Entailment)

\noindent \tb{T}: \emph{- Dziękuję - odpowiedział, czując, że policzek zaczyna go boleć.} \\
\tb{T}: [ENG] \emph{- 'Thank you', he replied, feeling his cheek begin to hurt.} 
\end{example}

In Example~\ref{ex:ex23} the epistemic verb is used, while in Example~\ref{ex:ex24} the main predicate is the perceptual verb. The former is non-factive and the latter is factive.

Further findings are that the models with inputs comprising text embeddings and linguistic features achieved slightly better results than those with only embedding inputs. Besides,
we can see that -- in our base feature model (1) -- some features make the most significant contribution to our model, i.e., if the verb is factive or non-factive (see Table~\ref{tab:feature-models} and Figure~\ref{fig:feature-importance}).
However, the indication of verb tense  (see Figure~\ref{fig:feature-importance}) as relatively important for our ML tasks, i.e. ECN classification, appears to be misleading in the light of linguistic data and requires further analysis. It seems that we are dealing here with spurious correlations rather than with lexical rules of language connecting the verb tense with ECN relations. Deeper linguistic analysis would be advisable here, however, because the relation between the grammatical tense of the verb and the ECN relations may be the result of pragmatic rules and prosodic properties of the utterances.
We hypothesize that these are spurious correlations in our dataset because, indeed, present or past tense co-occur more often with a particular class of ECN in our dataset.\footnote{Other names for this issue: \emph{annotation artifacts} \cite{gururangan2018annotation}, \emph{dataset bias} \cite{he2019unlearn}, \emph{group shift} \cite{oren2019distributionally}. For the problem of spurious correlations in the context of NLI see, e.g. \cite{dasgupta2018evaluating}; \cite{mccoy2019right}); \cite{10.1162/tacl_a_00335}.}

\section{Conclusions}
\label{sec:conclusions}

Machine learning solutions often act as black boxes that often feed on biases in training datasets. Therefore, it is not enough to use larger machines and larger corpora. Methodological reflection on the data gathering process, a precise definition of targeted tasks, and data quality evaluation are also essential.

In this study, we stated a review of the opposition factivity -- non-factivity in the context of predicting ECN relations. The dataset representing this phenomenon was gathered and analyzed. Then ML models were  trained based on this dataset.

Thus, we presented benchmarks with BERT-based models and models utilizing prepared linguistic features. They are even better than the performance of test annotators. However, a few issues remain unresolved in this task, i.e. utterances with a pragmatic foundation. Other issues to examine are potential spurious correlations (e.g. influence of the verb tense on the model results) -- further, deeper analysis of the models and their interpreting.  Our results indicate the need for a dataset that focuses on these kinds of cases.

\section*{Acknowledgements}
We want to thank Przemysław Biecek and Szymon Maksymiuk for their work on another NLI dataset and valuable remarks on conducting experiments and interpretability approaches. We also want to thank Karol Saputa, who implemented preliminary source code for the machine learning models we reused and redesigned in our experiments. Also, we are grateful for many students from the Faculty of Mathematics and Information Science at Warsaw University of Technology, working under Anna Wróblewska's guidance in Natural Language Processing course. They performed experiments on similar datasets and thus influenced our further research.

\bibliographystyle{unsrtnat}
\bibliography{references}  

\newpage

\appendix
\section{Examples From Our Dataset}
\label{sec:sup:examples}

In the following, there are a few examples from the \dataset dataset, and their descriptions. 

\begin{example}\label{ex:ex30} 
  
  \noindent \tb{T}:  [ENG] If a priest refuses Mass for a deceased person or a funeral because he received too little money, knowing that the payer is very poor, it is of course not right, but it is another matter. \\
  \tb{H}: The payer is very poor.

 \tb{T}: [PL] Jeżeli ksiądz odmówi mszy za zmarłego, lub pogrzebu z powodu zbyt niskiej zapłaty wiedząc, że proszący jest bardzo biedny to rzeczywiście nie jest w porządku, ale to inna sprawa. 
\\
  \tb{H}: Proszący jest bardzo biedny.
\\
Sentence type -- \textit{conditional} \\ 
Verb labels: \textit{wiedzieć, że / know that}, present, epistemic, factive, negation does not occur
\\
Complement tense -- \textit{present} \\ GOLD -- Neutral 
\end{example}

\textbf{Example~\ref{ex:ex30}}: Despite the factive verb \textit{wiedzieć, że}/\textit{know that}, GOLD label is neutral. This is because the whole utterance is conditional. An additional feature not included in the example is that the whole sentence does not refer to specific objects but is general.

\begin{example}\label{ex:ex31} 
  
  \noindent \tb{T}: [ENG] You never expected to hear that from me, did you? \\
  \tb{H}: You heard it from me.

 \tb{T}: [PL] Nie spodziewałeś się, że kiedykolwiek to ode mnie usłyszysz, co?
\\
\tb{H}: [PL] Kiedykolwiek to ode mnie usłyszysz.
\\
Sentence type -- \textit{interrogative}\\
Verb labels: \textit{spodziewać się, że / expect that}, past, epistemic, non-factive, negation occurs\\
Complement tense -- future \\
GOLD -- Entailment 
\end{example}

\textbf{Example~\ref{ex:ex31}}: Despite the non-factive verb \textit{spodziewać się, że}/\textit{expect that}, GOLD label is entailment. In this pair, non-lexical mechanisms are the basis of the entailment relation. Proper judgment of this example requires consideration of the prosodic structure of T's utterance.

It is worth noting that the sentence \textbf{H} is incorrectly written -- strictly speaking, it should be "\textbf{H'}:\textit{You have heard it from me.}/ \textit{Usłyszałeś to ode mnie}". So it is since \textbf{H} sentences were extracted semi-automatically. However, we did not want to change the linguistic features of the complement. The annotators were informed that in such situations, they should take into account not the \textbf{H} sentence, but its proper form -- in the above case, it is \textbf{H'}. From the perspective of bilingualism of the set, it is also vital that the information provided by the expression "\textit{never}" forms part of the main clause. In the Polish language, this content conveys the expression "\textit{kiedykolwiek}" and is part of the complement clause.

\begin{example}\label{ex:ex32} 
  
  \noindent \tb{T}: [ENG] maybe he was afraid that I would spill the beans... \\
\tb{H}: [ENG] I would spill the beans. \\

\tb{T}: [PL] może się bał że się wygadam... \\
\tb{H}: Się wygadam.
\\
Sentence type -- \textit{indicative} \\ Verb labels: \textit{bać się, że / afraid that}, past, emotive, non-factive, negation does not occur \\ GOLD -- ? 
\end{example}

\textbf{Example~\ref{ex:ex32}}: Example in which linguists decided to label a "?".\footnote{These utterances were removed from the dataset for training and testing our benchmarks.} Whether the state of affairs reflected by the complement clause was realized, it belongs to the common knowledge of the interlocutors. Without context, we are not able to say whether the sender spilled the beans or not. It is also worth noting that in the English translation, the modal verb is present. This element is absent in the Polish complement clause. We can also see that the lack of context does not make it possible to determine the \textbf{H} sentence.

\begin{example}\label{ex:ex33} 
  
  \noindent \tb{T}: [ENG] And that's why I made no effort to remind anyone of myself, I thought nobody here would remember me.
\tb{H}: nobody here would remember me.

\tb{T}: [PL] I dlatego nie starałem się przypomnieć, myślałem, że nikt tu o mnie nie pamięta. \\
\tb{H}: nikt tu o mnie nie pamięta
\\
Sentence type -- \textit{indicative} \\
Verb labels: \textit{myśleć, że / think that}, past, epistemic, non-factive, negation does not occur \\
GOLD -- Contradiction 
\end{example}

\textbf{Example~\ref{ex:ex33}}: The main verb is non-factive, and the relation between the whole sentence and its complement is a contradiction. The grounding of this relation has a pragmatic nature.

\section{Polish-English Translation}
\label{sec:sup:translation}

Our dataset is two-lingual. We translated its original Polish version into English. In the following, we described methodological challenges and doubts related to the creation of the second language version and the solutions we have adopted. 

We translated the whole dataset into English.
First, we used the deepL translator\footnote{https://www.deepl.com/translator},
then a professional translator corrected
the automatic translation. The human translator
acted following the guidelines: (a) not to change the structure
'[verb] {"to"|"that"} [complement clause]',
provided the sentence in English remained correct
and natural, (b) to keep various types of
linguistic errors in translation.

We believe that the decision on whether the translator
knows how to use the dataset is important
from the methodological point of view. Therefore,
we decided to inform the translator that it is crucial
for us that the translated sentence retains its
set of logical consequences, especially the relation
between the whole sentence and its complement clause. However, we did not provide the translator
with a GOLD column (annotations of specialist linguists).
The translator was aware that, in her task,
this aspect is essential. On the other hand, during
her work on each sentence, she had to assess the
Polish relation and try to keep it in translation.

The English version differs from the Polish in
several important issues. Each Polish sentence
contains a complementizer \emph{że}/\emph{that}. In English,
we can observe more complementizers, especially
\emph{that} \emph{to} and other, e.g., \emph{about}, \emph{for}, \emph{into}. There are also sentences without a complementizer. In
Polish, a complementizer cannot be elliptical, in
contrast to English (e.g., \emph{Nie
planowali, że będą mieli wielką, babską rodzinę}. /\emph{They
didn't plan they will have a big, girl family})
In some English sentences, an adjective, a noun, or a verb phrase has appeared instead of a verb, e.g., \emph{The English will appear in a weakened line-up for these meetings}. (in Polish: \emph{okazuje się, że Anglicy...})

It happens that depending on the sentence, the
Polish verb has more than one English equivalent,
e.g. \emph{cieszyć się} - \emph{glad that} or \emph{happy to}; \emph{realize that}
- \emph{zdawać sobie sprawę} or \emph{zrozumieć}. (In
the dictionaries \emph{zrozumieć} is closest to \emph{understand}).
For this reason, the frequency of verbs is different in respective sets.
Different language versions also pose problems related to verb signatures. First of all, the signatures developed by us are for Polish verbs. Therefore, we do not know how many pairs <V(pl); V(eng)> there
are, where verbs have identical signatures (factive or non-factive). Secondly, a verb in language L1 may not have its equivalent in language L2 and vice versa.

\clearpage 

\section{Test Annotations by Non-Experts}
\label{sec:sup:annotators}

Table~\ref{tab:non-expert-annotations} shows examples of annotations made by non-experts in our study.

\begin{table*}[!htb]
\caption{Annotation examples for non-experts. Note: "Annot." indicate non-expert annotations.\label{tab:non-expert-annotations}}
\centering
\begin{tabular}{p{0.8cm}p{12.5cm}}
\toprule
Label name & Value \\
\midrule
T & [ENG] \emph{I know the cold degrades the mind and makes it sluggish.} 

[PL] \emph{Wiem, że zimno degraduje umysł i wiedzie go do ospałości.}
\\
H & [ENG] \emph{The cold degrades the mind and makes it sluggish.} 

[PL] \emph{Zimno degraduje umysł i wiedzie go do ospałości}\\
Task & GOLD -- E,  Annot. -- E E E E \\
\midrule
T & [ENG] \emph{Statists believe that people can be demanding towards the state.} 

[PL] \emph{Etatyści wierzą, że ludzie mogą wymagać od państwa. 
}
\\
H & [ENG] \emph{People can be demanding towards the state.} 

[PL] \emph{Ludzie mogą wymagać od państwa.}\\
Task & GOLD -- N, Annot. -- N N N N \\
\midrule
T & [ENG] \emph{I thought you were a bachelor.} 

[PL] \emph{Myślałam, że jesteś kawalerem.}
\\
H & [ENG] \emph{You were a bachelor} 

[PL] \emph{Jesteś kawalerem.}\\
Task & GOLD -- N, Annot. -- C C N N \\
\midrule
T & [ENG] \emph{I imagined that if there was guilt, then there was punishment.} 

[PL] \emph{Wyobrażałem sobie, że jak jest wina, to jest i kara.}
\\
H & [ENG] \emph{If there was guilt, then there was punishment.} 

[PL] \emph{Jak jest wina, to jest i kara.}\\
Task & GOLD -- C, Annot. -- N C N C \\
\midrule
T & [ENG] \emph{She may have ordered hastily, but she really hoped that the stadium would remain a place of trade, not sport.} 

[PL] \emph{Być może zamawiała pochopnie, ale naprawdę liczyła, że stadion pozostanie miejscem handlu, nie sportu.}
\\
H & [ENG] \emph{The stadium would remain a place of trade, not sport.} 

[PL] \emph{Stadion pozostanie miejscem handlu.}\\
Task & GOLD -- C, Annot. -- N C N N \\
\midrule
T & [ENG] \emph{I was wondering when you could talk about a woman's life making sense...} 

[PL] \emph{Zastanawiałam się, kiedy można mówić o tym, że życie kobiety miało sens... } \\
H & [ENG] \emph{A woman's life making sense...} 

[ENG] \emph{Życie kobiety miało sens...}\\
Task & GOLD -- N, Annot. -- N ? N ? \\
\midrule
T & [ENG] \emph{I mean, Mrs. W. didn't say she was kicked out of the house.} 

[PL] \emph{Przecież pani W. nie powiedziała, że została wyrzucona z domu.}
\\
H & [ENG] \emph{She was kicked out of the house.} 

[PL] \emph{Została wyrzucona z domu.}\\
Task & GOLD -- E, Annot. -- N N C C \\
\midrule
\midrule
T & [ENG] \emph{Szerucki wiped his wet cheeks with a frayed sleeve, walked in, closed the door behind him and felt that something was wrong.
} 

[PL] \emph{Szerucki przetarł wystrzępionym rękawem zroszone policzki, wszedł, zamknął za sobą drzwi i poczuł, że jest niedobrze.
}
\\
H & [ENG] \emph{Something was wrong.} 

[PL] \emph{Jest niedobrze.}\\
Task & GOLD -- E, Annot. -- N N E E \\
\midrule
\bottomrule
\end{tabular}
\end{table*}

\end{document}